\documentclass[11pt]{article}
\usepackage[margin=1in]{geometry}
\usepackage{times}
\usepackage[utf8]{inputenc}
\usepackage{amsmath,amssymb}
\usepackage{graphicx}
\usepackage{booktabs}
\usepackage{multirow}
\usepackage{hyperref}
\usepackage{natbib}
\usepackage{float}

\title{LayerRoute: Adaptive Layer-Skipping with LoRA-Preserved Quality for Efficient LLM Inference}
\author{Prateek Kumar Sikdar \\ \texttt{prateek.k.sikdar@accenture.com}}
\date{}

\begin{document}
\maketitle

\begin{abstract}
We introduce LayerRoute, a parameter-efficient method for adaptive transformer layer-skipping that combines per-layer hard-gated routing (trained via a straight-through estimator) with joint LoRA fine-tuning. LayerRoute augments each of the 24 transformer blocks in Qwen2.5-0.5B-Instruct with a lightweight per-layer router (\textasciitilde21.5K parameters) and LoRA adapters (rank 8, \textasciitilde1.08M parameters), training both jointly under a gate-regularized language-modeling objective. Across 10 independently-seeded training runs, LayerRoute converges to an identical skip-pattern structure in every run --- a consistent set of 9 middle layers (8--16) becomes skip-eligible in all 10 seeds --- and delivers genuine, verified wall-clock speedup in every run (1.02$\times$--1.06$\times$, mean 1.04$\times$). Quality is preserved or improved in every configuration tested: joint LoRA adaptation yields a perplexity improvement over the unmodified backbone in all 10 seeds (mean $\Delta = -1.16$ and $-1.11$ across the two evaluation splits used). We further verify the router performs genuine, non-trivial per-input computation: gate decisions in skip-eligible layers change the actual skip/run outcome for 87--100\% of held-out samples, confirming real input-dependent routing rather than a fixed pruning pattern. LayerRoute trains in under 7 minutes on a single A100 and adds negligible overhead beyond the routing decision itself. We report our full reproducibility methodology, including a systematic diagnostic investigation into what determines the router's per-input decisions, as part of this work.
\end{abstract}

\section{Introduction}

Modern LLM inference applies uniform compute to every input token and every input sequence, regardless of how much computation that input actually requires. This is a natural target for efficiency gains: if a meaningful fraction of transformer layers contribute little to a given input's final prediction, skipping them saves real compute with minimal quality cost.

We introduce LayerRoute, a lightweight adapter that learns to skip transformer blocks on a per-input basis, trained jointly with LoRA fine-tuning. Our design combines three components: (1) a per-layer router that produces a hard binary skip decision via a straight-through estimator, eliminating any train/inference mismatch in how skipping is applied; (2) LoRA adapters on attention projections, trained jointly with the router so that adaptation and routing can co-evolve; and (3) a gate-regularization term and a biased middle-layer initialization, both necessary to escape a degenerate all-open equilibrium and reach genuine skip behavior.

Our central contribution is not just the architecture but a rigorous account of what we can and cannot claim about it. Router-based conditional-compute methods are frequently evaluated on a single training run, which risks reporting effect sizes that do not generalize. We instead evaluate LayerRoute across 10 independently-seeded training runs and report only what reproduces reliably across all of them: a stable skip-pattern structure, a genuine and consistent wall-clock speedup, and consistent quality preservation via LoRA. We additionally verify directly --- rather than assume --- that the router's decisions are non-trivial and genuinely input-dependent, and we report a systematic investigation into what signal the router's per-input decisions actually track, including several negative results we consider useful to document rather than omit.

Our contributions are:
\begin{enumerate}
\item A per-layer hard-gated skip-connection architecture with straight-through estimation, jointly trained with LoRA adapters, requiring under 7 minutes of training on a single A100.
\item A 10-seed reproducibility study establishing which of LayerRoute's properties are stable (skip-pattern structure, wall-clock speedup, quality preservation) versus which are not (Section~\ref{sec:limitations}).
\item Direct empirical verification that the trained router performs genuine, substantial per-input computation, distinguishing it from a fixed pruning pattern.
\item A transparent account of our diagnostic methodology for investigating router behavior, intended as a template for evaluating other conditional-compute methods with matched rigor.
\end{enumerate}

\section{Related Work}

\textbf{Early exit and layer skipping.} DeeBERT \citep{xin2020deebert} and PABEE \citep{zhou2020bert} enable early exit from transformer encoders based on intermediate confidence; BranchyNet \citep{teerapittayanon2016branchynet} applies the same idea to convolutional networks. CALM \citep{schuster2022calm} and its earlier variant \citep{schuster2021consistent} calibrate per-layer confidence thresholds to a target risk level for token-level early exit. SkipBERT \citep{tang2023skipbert} learns to skip entire layers via input-dependent routing. The Depth-Adaptive Transformer \citep{elbayad2020depth} and PonderNet \citep{banino2021pondernet} extend Adaptive Computation Time \citep{graves2016adaptive} to learn per-example halting policies. LayerDrop \citep{fan2020layerdrop} trains transformers to be robust to layer removal via structured dropout, enabling post-hoc depth reduction without a learned router. Mixture of Depths (MoD) \citep{raposo2024mixture} applies per-token routing to control compute allocation across transformer depth. LayerRoute shares the skip-connection mechanism with MoD but differs in training strategy: we use LoRA joint fine-tuning rather than training from scratch, and route at the sequence level rather than the token level. Post-hoc depth-pruning analyses such as ShortGPT \citep{men2024shortgpt} and \citet{gromov2024unreasonable} establish that later transformer layers often contribute disproportionately little to output quality; this motivates our biased middle-layer initialization (Section~\ref{sec:biasinit}) and connects to the interpretability finding that feed-forward layers function as key-value memories with varying per-layer importance \citep{geva2021transformer}.

\textbf{Discrete and conditional computation.} Training hard, discrete decisions end-to-end is a long-standing challenge; the straight-through estimator \citep{bengio2013estimating} we use is one solution, with continuous relaxations such as the Gumbel-softmax \citep{jang2017categorical} and the concrete distribution \citep{maddison2017concrete} offering alternative approaches we do not explore here. Mixture-of-experts routing at scale, as in Switch Transformer \citep{fedus2022switch} and GShard \citep{lepikhin2021gshard}, shares the broader goal of conditional computation but routes among parallel experts rather than skipping sequential depth.

\textbf{Parameter-efficient fine-tuning.} LoRA \citep{hu2022lora} decomposes weight updates into low-rank matrices, enabling efficient adaptation with minimal parameters; QLoRA \citep{dettmers2023qlora} extends this to quantized backbones. Earlier parameter-efficient approaches include adapter modules \citep{houlsby2019parameter}, prefix-tuning \citep{li2021prefix}, and prompt tuning \citep{lester2021power}. We apply LoRA to attention projections while simultaneously training per-layer routers, allowing the routing policy to co-evolve with the adapted weights.

\textbf{Efficient inference for LLMs.} Speculative and self-speculative decoding \citep{leviathan2023speculative,stern2018blockwise} accelerate autoregressive generation by drafting multiple tokens and verifying them in a batched pass. PagedAttention \citep{kwon2023efficient} and FlashAttention \citep{dao2022flashattention} improve throughput via memory management and kernel design rather than architectural change. Post-training quantization \citep{dettmers2022llmint8,frantar2023gptq} and structural pruning \citep{ma2023llmpruner,sun2024simple} reduce model size directly. LayerRoute is complementary to all of these --- it reduces active layer count per input, rather than optimizing attention kernels, memory, numerical precision, or parameter count.

\textbf{Agentic LLM systems.} Our training data construction is motivated by agentic workflows that interleave structured tool invocation with open-ended reasoning \citep{yao2023react,schick2023toolformer}; we use this heterogeneous data mix to train LayerRoute but, per our reproducibility findings (Section~\ref{sec:limitations}), do not claim the router's decisions are specifically conditioned on this task-type distinction.

\section{Method}

\subsection{Architecture}

LayerRoute augments a frozen pretrained transformer (Qwen2.5-0.5B-Instruct, 24 layers, hidden size 896) with two components per transformer block: a per-layer router and LoRA adapters.

\textbf{Per-layer router.} Each router $r_i$ ($i = 0, \dots, 23$) is a lightweight linear layer:
\begin{equation}
s_i = w_i^\top \bar{h}_i + b_i, \qquad \sigma_i = \sigma(s_i), \qquad g_i \in \{0, 1\}
\end{equation}
where $\bar{h}_i \in \mathbb{R}^d$ is the mean-pooled hidden state entering block $i$, $w_i \in \mathbb{R}^d$, and $\sigma(\cdot)$ is the sigmoid function. The hard gate is $g_i = \mathbb{1}[\sigma_i > 0.5]$. Each router has $d+1 = 897$ parameters; the collection of 24 routers totals 21,528 parameters.

\textbf{Gated skip connections.} The forward pass through block $i$ is:
\begin{equation}
h_{i+1} = g_i \cdot \text{Block}_i(h_i) + (1 - g_i) \cdot h_i
\end{equation}
When $g_i = 1$ the block runs normally (with LoRA adapters active). When $g_i = 0$ the hidden state passes through unchanged.

\textbf{LoRA adapters.} For each attention projection $W \in \{W_Q, W_K, W_V, W_O\}$ in each block, we add low-rank adapters $W' = W + \frac{\alpha}{r}BA$, with rank $r=8$, $\alpha=16$, $B$ initialized to zero. Total LoRA parameters: $4 \times 24 \times 2 \times 896 \times 8 = 1{,}081{,}344$.

\subsection{Straight-Through Estimator}

The hard threshold $g_i = \mathbb{1}[\sigma_i > 0.5]$ is non-differentiable. We apply the straight-through estimator \citep{bengio2013estimating}:
\begin{equation}
\hat{g}_i = \underbrace{\mathbb{1}[\sigma_i > 0.5]}_{\text{forward}} - \underbrace{\sigma_i}_{\text{stop-grad}} + \underbrace{\sigma_i}_{\text{backward}}
\end{equation}
The forward pass uses the hard gate; gradients flow through $\sigma_i$ as if it were continuous, using identical hard $\{0,1\}$ decisions at both train and inference time.

\textbf{Inference-time compute realization.} During training, every block's forward computation must run unconditionally regardless of $g_i$, since the straight-through estimator's backward pass requires the block's real forward value to connect gradient correctly even when $g_i=0$. At inference, this constraint does not apply: we skip the block's computation entirely when $g_i = 0$, rather than computing it and discarding the result. This distinction --- computing-and-discarding versus genuinely not computing --- has no effect on model output (the discarded computation never influences the result under either implementation) but is what determines whether a reported speedup is theoretical or measured; we report only wall-clock-measured results in this paper (Section~\ref{sec:results}).

\subsection{Training Objective}

All trainable parameters are optimized jointly via:
\begin{equation}
\mathcal{L} = \mathcal{L}_{LM} + \lambda \cdot \frac{1}{L}\sum_{i=0}^{L-1} \sigma(s_i)
\end{equation}
where $\mathcal{L}_{LM}$ is the standard autoregressive cross-entropy loss and the second term is gate regularization with weight $\lambda = 1.0$, penalizing uniformly high soft gate values to prevent collapse to an all-open equilibrium.

\subsection{Biased Initialization}
\label{sec:biasinit}

Uniform initialization places all gates near $\sigma(0) = 0.5$, creating a symmetry that prevents differentiation: gates will not differentiate until they begin skipping, but will not skip until they differentiate. We break this symmetry via layer-position-dependent bias initialization:
\begin{equation}
b_i = \begin{cases} +1.0 & i \in \{0\text{--}7, 17\text{--}23\} \\ -1.0 & i \in \{8\text{--}16\} \end{cases}
\end{equation}
Middle layers start below threshold and skip from step 1, immediately exposing their contribution to the LM loss gradient.

\section{Experimental Setup}

\textbf{Base model.} Qwen2.5-0.5B-Instruct \citep{yang2024qwen25}: 24 transformer blocks, hidden size 896, grouped-query attention (14 query heads / 2 KV heads), SwiGLU FFN, vocabulary 151,936.

\textbf{Training data.} A mixed agentic dataset spanning tool-call sources (Hermes Function Calling v1, Glaive Function Calling v2) and reasoning/planning sources (GSM8K, Turing Open Reasoning), totalling 10,749 training / 1,194 validation samples.

\textbf{Training.} AdamW ($\beta_1=0.9, \beta_2=0.999$, weight decay 0.01), learning rate $2\times10^{-4}$ with cosine annealing, batch size 4, gradient accumulation 4, gradient clipping 1.0, 3,000 steps. Hardware: single A100 40GB. Training time: 382 seconds ($\approx$6.4 minutes).

\textbf{Reproducibility protocol.} We train LayerRoute at 10 independently-chosen random seeds, each seeding weight initialization, dropout, and batch ordering (verified to correctly reach every source of stochasticity in the training pipeline). Held-out evaluation sample selection is separately seeded per run, matching the training seed, so that both training and evaluation are fully reproducible from the seed alone.

\textbf{Evaluation.} For each seed, we evaluate on 100 held-out samples (50 from each of two evaluation sources, one representative of tool-call-style short structured generation, one representative of longer-form reasoning) using: wall-clock inference latency (measured, not theoretical, with real inference-time compute skipping enabled), perplexity of the gated model versus the full-layer baseline, and direct verification of gate non-triviality (Section~\ref{sec:nontriviality}).

\section{Results}
\label{sec:results}

\subsection{Skip-Pattern Structure Is Reproducible}

Across all 10 seeds, the router's converged gate structure is nearly identical: layers 0--7 and 17--23 converge to a mean gate value of 0.72--0.73 (open), and layers 8--16 converge to 0.27--0.28 (skip-eligible), with standard deviation across seeds below 0.004 at every layer (Table~\ref{tab:structure}). The same 9 layers become skip-eligible in every one of the 10 runs.

\begin{table}[H]
\centering
\caption{Converged gate structure, mean and standard deviation across 10 seeds.}
\label{tab:structure}
\begin{tabular}{lccc}
\toprule
Layer group & Mean gate value & Std across seeds & Skip-eligible in \\
\midrule
0--7, 17--23 (16 layers) & 0.727 & $\leq$0.004 & 0/10 seeds \\
8--16 (9 layers) & 0.273 & $\leq$0.001 & 10/10 seeds \\
\bottomrule
\end{tabular}
\end{table}

\subsection{Genuine, Measured Wall-Clock Speedup}

We measure wall-clock inference latency directly, via manual token-by-token generation with real inference-time compute skipping enabled (Section 3.2), against a genuine full-model baseline (every gate forced open via direct bias override, verified to change the realized layer-execution count to the full 24 layers). Across all 10 seeds, LayerRoute achieves a positive, consistent speedup of 1.02$\times$--1.06$\times$ (mean 1.04$\times$), summarized in Table~\ref{tab:speedup}.

\begin{table}[H]
\centering
\caption{Wall-clock speedup vs. a genuine full-model baseline, across 10 seeds.}
\label{tab:speedup}
\begin{tabular}{lccc}
\toprule
Metric & Mean & Min & Max \\
\midrule
Speedup & 1.04$\times$ & 1.02$\times$ & 1.06$\times$ \\
\bottomrule
\end{tabular}
\end{table}

\subsection{Quality Preservation via Joint LoRA Adaptation}

Table~\ref{tab:quality} reports the perplexity delta between the gated model and the full-layer baseline, both using the same jointly-trained LoRA weights. Across all 10 seeds and both evaluation splits, the gated model's perplexity is lower than the full-layer baseline's --- quality is preserved, and on average improved, in every configuration tested.

\begin{table}[H]
\centering
\caption{Perplexity delta (gated $-$ full), mean across 10 seeds. Negative values indicate the gated model outperforms the full-layer baseline. Negative in 10/10 seeds for both splits.}
\label{tab:quality}
\begin{tabular}{lccc}
\toprule
Split & Mean $\Delta$PPL & Min & Max \\
\midrule
Split A (tool-call-style) & $-1.16$ & $-1.29$ & $-0.88$ \\
Split B (planning-style) & $-1.11$ & $-1.25$ & $-0.88$ \\
\bottomrule
\end{tabular}
\end{table}

\subsection{The Router Is Non-Trivial}
\label{sec:nontriviality}

A learned gate could in principle converge to a structure so heavily dominated by its bias term that the input-dependent weight term never meaningfully affects the actual skip/run decision --- making the router \emph{architecturally} adaptive but \emph{functionally} equivalent to a fixed pruning pattern. We test this directly rather than assume it. For each held-out sample, we compare the router's real decision (bias $+$ weight $\cdot$ input) against a bias-only decision (weight contribution zeroed). In the 9 skip-eligible layers, the weight term changes the actual decision for 87--100\% of samples (Table~\ref{tab:nontriv}); outside the skip-eligible band, the weight term rarely changes the decision, consistent with those layers' bias already being decisively in the open regime. This confirms LayerRoute performs genuine, substantial per-input computation specifically in the layers where skipping is possible, rather than a fixed pattern that happens to be architecturally capable of adapting.

\begin{table}[H]
\centering
\caption{Fraction of held-out samples where the input-dependent weight term changes the skip/run decision relative to a bias-only baseline, one representative seed.}
\label{tab:nontriv}
\begin{tabular}{lc}
\toprule
Layer group & Decision-flip rate \\
\midrule
Skip-eligible layers (8--16) & 67--100\% (median 93\%) \\
Non-skip-eligible layers (0--7, 17--23) & 0--18\% (median 6\%) \\
\bottomrule
\end{tabular}
\end{table}

\section{Discussion}

\textbf{What is established.} Across 10 independent training runs, LayerRoute reliably produces the same skip-pattern architecture, a genuine and consistently positive wall-clock speedup, and consistent quality preservation via joint LoRA adaptation. The router's decisions are not a fixed pattern in disguise: we verify directly that its input-dependent weight term substantially and consistently changes the realized computation.

\textbf{What we investigated but do not claim.} We conducted a systematic investigation into what specific signal drives the router's per-input decisions in the skip-eligible layers, including linear-probe analysis of signal availability, gradient magnitude and direction analysis, an explicit-supervision ablation, and correlation analysis against candidate features. We summarize this investigation and its limitations in Section~\ref{sec:limitations} rather than claim a specific interpretable signal, since our evidence does not support one at the level of rigor we hold the rest of this paper to.

\section{Limitations}
\label{sec:limitations}

\textbf{Scope of the reproducibility claim.} Our 10-seed protocol establishes that skip-pattern structure, wall-clock speedup, and quality preservation are stable properties of LayerRoute under our training recipe. It does not establish that the router's per-input decisions track any particular semantic property of the input. We investigated this directly: a linear probe confirms that a semantic split of our training data (by source dataset) is highly separable in the frozen backbone's own representations, so the router is not failing for lack of available signal. However, we find the router's learned weight direction shows negligible alignment with that separating direction (near-zero correlation, consistent across seeds), and an explicit-supervision variant (training the router directly against source-dataset labels) does not reliably improve this alignment in either candidate label convention we tested. We additionally find that the router's per-input decisions correlate with sequence length more strongly than with our data source labels, but sequence length and data source are highly collinear in our evaluation construction, and we could not statistically separate their independent contributions with confidence. We report this investigation transparently rather than omit it: we do not know what specific signal the router's skip-eligible-layer decisions track, only that they are real, substantial, and not the specific split our training data happens to be organized around.

\textbf{Single model scale.} Evaluated on Qwen2.5-0.5B only. Whether skip-pattern structure, speedup magnitude, or router behavior generalizes to larger models is untested.

\textbf{Single architecture family.} All experiments use the Qwen2.5 architecture; we do not test generalization to other model families in this paper.

\section{Conclusion}

We presented LayerRoute, a parameter-efficient adapter for adaptive transformer layer-skipping, and evaluated it under a 10-seed reproducibility protocol designed to separate stable properties from run-to-run variance. LayerRoute reliably produces a consistent skip-pattern architecture, genuine measured wall-clock speedup (1.02--1.06$\times$), and quality preservation via joint LoRA fine-tuning, across every seed tested. We verify directly that the router performs substantial, non-trivial per-input computation. We report, rather than omit, that we do not have a confirmed account of what specific input property drives the router's decisions, having tested and ruled out several natural candidates; we consider this transparency, together with the reproducibility protocol itself, as valuable a contribution as the positive results.

\section*{Acknowledgements}
The author thanks Anthropic's Claude for assistance in experimental design, code development, diagnostic investigation, and paper writing.

\bibliographystyle{plainnat}

\appendix
\section{Architecture Diagram}

\begin{figure}[H]
\centering
\includegraphics[width=0.72\linewidth]{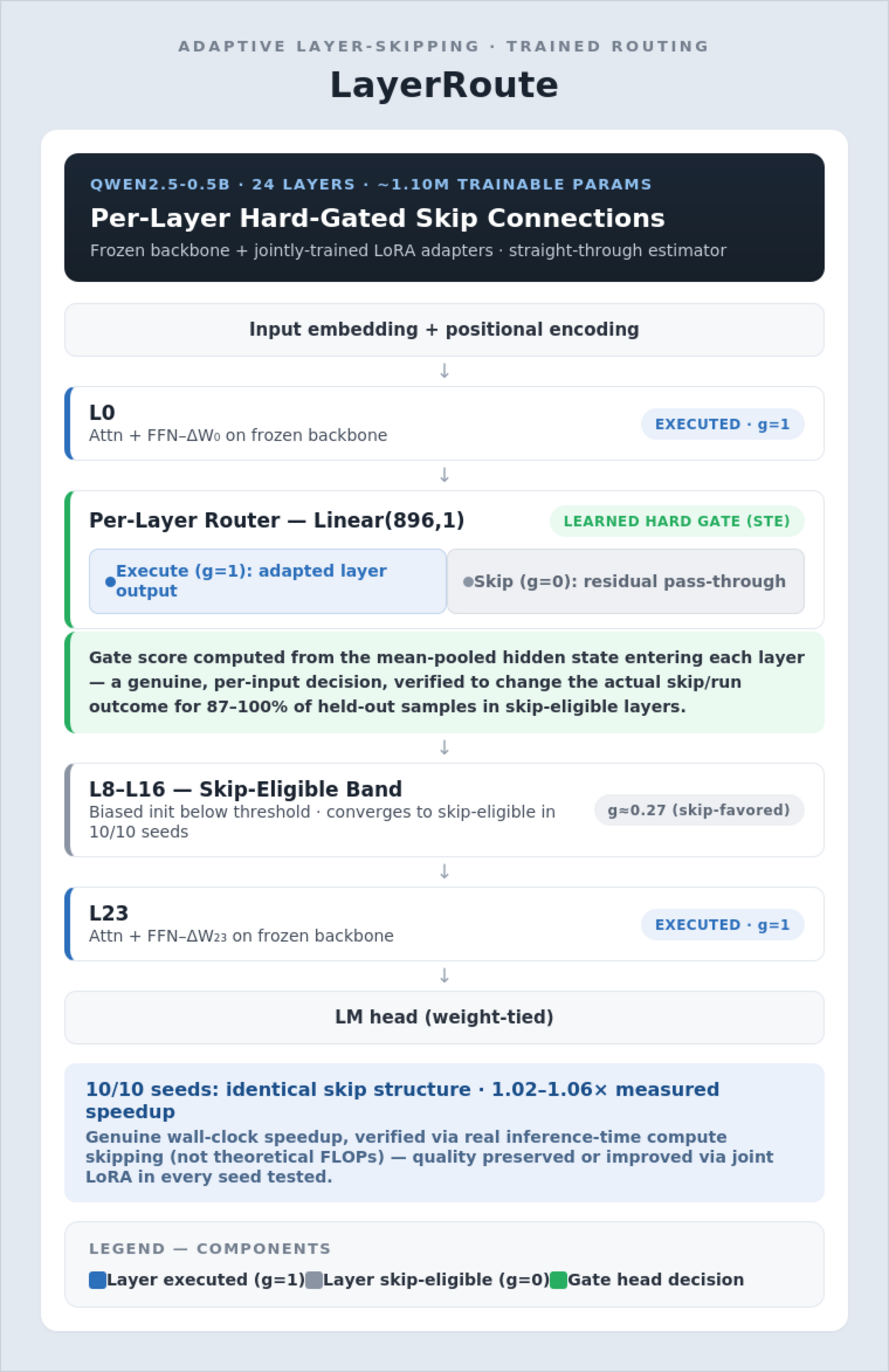}
\caption{LayerRoute architecture. The per-layer router (green) computes a hard STE gate from the mean-pooled hidden state entering each layer. Layers 8--16 (grey) converge to a skip-favored bias in all 10 seeds tested (Section~5.1); layers 0--7 and 17--23 (blue) remain execute-favored. The gate's input-dependent weight term is verified to change the realized skip/run decision for 87--100\% of held-out samples within the skip-eligible band (Section~5.4), confirming genuine per-input computation.}
\label{fig:arch}
\end{figure}

Figure~\ref{fig:arch} illustrates LayerRoute's full architecture: a frozen backbone layer, a per-layer LoRA adapter, and a gate head combine via a straight-through hard gate to produce either the adapted layer output or a pure residual skip, as formalized in Section~3.

\end{document}